\documentclass{article}
\usepackage{style/spconf}

\usepackage{verbatim}
\usepackage{xspace}
\usepackage{setspace}

\usepackage{booktabs}

\usepackage{multicol}

\usepackage{caption}
\usepackage{subcaption}
\usepackage{fancyhdr}
\usepackage{multirow}
\usepackage{enumitem}
\setlist{nosep, leftmargin=14pt}
\usepackage{url}            

\usepackage{graphicx}
\usepackage[
usenames,
dvipsnames,
hyperref]{xcolor}

\usepackage{amsmath}
\usepackage{amssymb}
\usepackage{amsfonts} 
\usepackage{bm}
\usepackage{dsfont}
\usepackage[tbtags]{mathtools}
\usepackage{nicefrac}

\usepackage{cite}     
\usepackage[pagebackref,breaklinks,colorlinks]{hyperref}
\usepackage[capitalize]{cleveref}

\crefname{section}{Sec.}{Secs.}
\Crefname{section}{Section}{Sections}
\Crefname{table}{Table}{Tables}
\crefname{table}{Tab.}{Tabs.}

\definecolor{LightGray}{rgb}{0.92,0.92,0.92}

\newcommand{\rom}[1]{\uppercase\expandafter{\romannumeral #1\relax}}

\newcommand{\mmu}[1]{\mbox{#1 {\textmu}m}\xspace}

\newcommand{\mpx}[1]{\mbox{#1 pixels}\xspace}

\newcommand{\mpli}{\mbox{3D-PLI }}

\fancypagestyle{specialfooter}{%
  \fancyhf{}
  
  \fancyfoot[C]{
    \fontsize{9}{11}\selectfont\textbf{\textcopyright 2024 IEEE. Personal use of this material is permitted.  Permission from IEEE must be obtained for all other uses, in any current or future media, including reprinting/republishing this material for advertising or promotional purposes, creating new collective works, for resale or redistribution to servers or lists, or reuse of any copyrighted component of this work in other works.}
  }
}
\title{Analyzing Regional Organization of the Human Hippocampus in 3D-PLI Using Contrastive Learning and Geometric Unfolding}

\name{
  Alexander Oberstrass$^{1,2}$ ~ Jordan DeKraker$^{3}$ ~ Nicola Palomero-Gallagher$^{1,4}$ ~  Sascha E. A. Muenzing$^{1}$
}{
  Alan C. Evans$^{3}$ ~ Markus Axer$^{1,5}$ ~ Katrin Amunts$^{1,4}$ ~ Timo Dickscheid$^{1,2,6}$
}
\address{
  $^{1}$ Institute of Neuroscience and Medicine (INM-1), Research Centre Jülich, Germany \\
  $^{2}$ Helmholtz AI, Research Centre Jülich, Germany \\
  $^{3}$ Montreal Neurological Institute and Hospital, McGill University, Montreal, Canada \\
  $^{4}$ Cécile \& Oskar Vogt Institute of Brain Research, University Hospital Düsseldorf, Germany \\
  $^{5}$ Department of Physics, University of Wuppertal, Germany \\
  $^{6}$ Institute of Computer Science, Heinrich-Heine-University Düsseldorf, Germany \\
}

\begin{document}

\thispagestyle{specialfooter}
\maketitle

\begin{abstract}
Understanding the cortical organization of the human brain requires interpretable descriptors for distinct structural and functional imaging data.
3D polarized light imaging \mbox{(3D-PLI)} is an imaging modality for visualizing fiber architecture in postmortem brains with high resolution that also captures the presence of cell bodies, for example, to identify hippocampal subfields.
The rich texture in \mpli images, however, makes this modality particularly difficult to analyze and best practices for characterizing architectonic patterns still need to be established.
In this work, we demonstrate a novel method to analyze the regional organization of the human hippocampus in \mpli by combining recent advances in unfolding methods with deep texture features obtained using a self-supervised contrastive learning approach.
We identify clusters in the representations that correspond well with classical descriptions of hippocampal subfields, lending validity to the developed methodology.
\end{abstract}
%

\section{Introduction}
\label{sec:intro}
A comprehensive understanding of human brain organization requires measuring its distinct structural and functional properties and making them accessible in the form of spatial maps and multimodal brain atlases \cite{amunts2015}.
Previous work on mapping the hippocampus, an archicortical structure crucial for learning and memory, emphasizes the importance of integrating multimodal data to confirm known brain regions identified by cytoarchitectonic criteria and reveal new subdivisions~\cite{palomero-gallagher2020}.

To contribute to the joint effort in multimodal mapping, we aim to characterize the hippocampal architecture based on three-dimensional polarized light imaging (3D-PLI)~\cite{axer2011}.
\mpli offers a staining-free microscopic imaging technique for whole postmortem brain sections that exclusively utilizes optical properties of the tissue.
It enables both the measurement of nerve fiber orientations with microscopic resolution, revealing fine-grained structures such as small fiber tracts, as well as the presence of cell bodies through lower light transmittance \cite{zeineh2017}.
The rich texture information provided by \mpli enables visual identification of hippocampal subfields~\cite{zeineh2017}.
This process, however, requires anatomical experience and is enormously time-consuming.
Since variations in \mpli texture are complex and challenging to quantify, observer-independent mapping, as performed for cell-body stained sections~\cite{schleicher1999}, has not yet been achieved and best practices for representing \mpli still need to be established.

In this work, we aim to broaden the interpretation of \mpli by automatically characterizing the regional organization of the pyramidal layer in the human hippocampus.
We build on deep texture features generated by a recently introduced self-supervised contrastive learning approach~\cite{oberstrass2024}.
To account for the highly folded morphology of the hippocampus, we build on unfolding methods \cite{dekraker2018,dekraker2020,dekraker2023} to sample features at different depths within the pyramidal layer and project them into a flat reference space.
To evaluate the distinctiveness of this approach, we compare identified clusters in the texture features with known subfields.
Our contributions are the following:
\vspace*{\fill}
\begin{itemize}
  \setlength\itemsep{1.4mm}
  \item We demonstrate a novel method to analyze the regional organization of the human hippocampus in \mpli by combining unfolding methods \cite{dekraker2018} with deep texture features by a self-supervised contrastive learning objective \cite{oberstrass2024}.
  \item We demonstrate that texture features learned by a cross-section sampling strategy \cite{oberstrass2024} to generate positive pairs in contrastive learning enable more stable clustering by hippocampal subfields compared to in-plane sampling and classical features derived from \mpli parameters.
  \item We perform an ablation study of individual input modalities and show that adding fiber orientations to the input for feature extraction improves correspondence of clusters to subfields over features from transmittance maps alone.
\end{itemize}

\section{Methods}
\label{sec:methods}

\begin{figure}[t!]
    \centering
    \includegraphics[width=\linewidth]{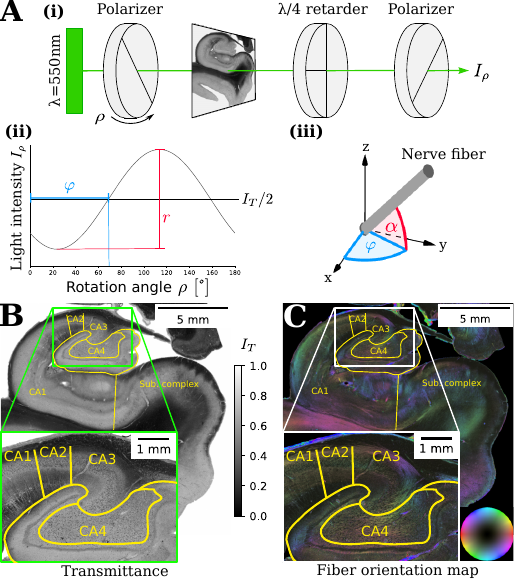}
    \caption{
      \mpli measurement process and example parameter maps for a brain section through a human hippocampus.
      (A) Setup of the polarizing microscope (PM). For rotations $\rho$ of a linear polarizer (i), an intensity profile is recorded for each pixel (ii), from which transmittance $I_T$, fiber direction $\varphi$ and inclination $\alpha$ (from retardation $r$) can be derived. $\varphi$ and $\alpha$ determine the 3D orientation of nerve fibers (iii).
      (B) Example transmittance map $I_T$ with hippocampal subfield labels CA1, CA2, CA3, CA4 and the Subicular complex.
      (C) Fiber orientation map (FOM) of the same section as in (B).
    }
    \label{fig:data}
\end{figure}

\textbf{\mpli hippocampus.}
In this work, we utilize a 3D reconstructed human hippocampus of an 87-year-old male \cite{beaujoin2018} measured with a polarizing microscope (PM) \cite{axer2011}.
The 3D volume is composed of 545 individual brain sections, each 26757 $\times$ 22734 pixels in size.
Damaged or missing sections are replaced by their nearest neighbor.
Truncated absolute anterior and posterior parts are omitted for analysis.

The measurement (\cref{fig:data}A) was carried out at \mmu{1.3} in-plane resolution on individual \mmu{60} thick brain sections.
From the measured light intensity values, \mpli parameter maps transmittance $I_T$ (\cref{fig:data}B), retardation $r$, and direction $\varphi$ (in-plane orientation) were derived.
A transmittance-weighted model \cite{menzel2022} was applied to derive fiber inclination $\alpha$ (out-of-plane orientation) from retardation values $r$.
The resulting three-dimensional orientation of nerve fibers was visualized in fiber orientation maps (FOM; \cref{fig:data}C).
FOMs encode direction $\varphi$ as hue and inclination $\alpha$ as saturation and value in HSV color space (darker areas mark higher inclination).

\begin{figure*}[h!t]
    \centering
    \includegraphics[width=\textwidth]{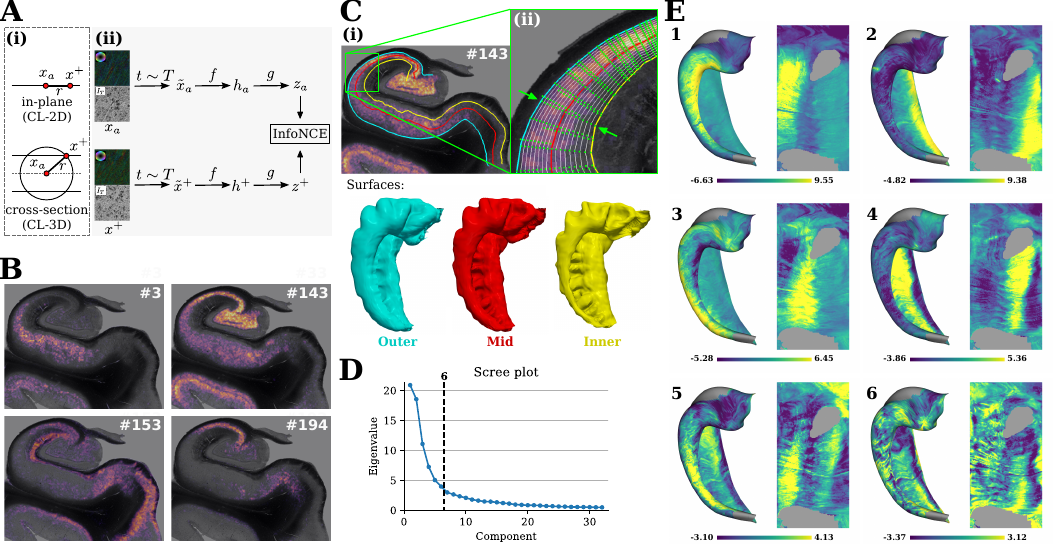}
    \caption{
        Overview of the deep texture feature extraction and unfolding.
        (A) Encoder models $f$ are trained by (i) sampling positive pairs at a fixed distance $r$ either in-plane (\mbox{CL-2D}) or across brain sections (\mbox{CL-3D}) in (ii) a contrastive learning framework.
        (B) Selected feature maps generated by \mbox{CL-3D}.
        A sliding window approach is used to generate feature maps for whole brain sections.
        Feature maps are overlaid with transmittance maps for reference.
        (C) Coronal view of surfaces extracted using \textit{HippUnfold}, overlaid on feature \#143 and transmittance (i).
        Deep texture features by \mbox{CL-3D} are sampled along interpolated vertices between geometrical inner and outer surfaces (green arrows; ii) and concatenated into single vectors.
        (D) Scree plot showing highest eigenvalues for the first 6 PCA components calculated for the concatenated texture features.
        (E) Unfolded projections of the concatenated features by \mbox{CL-3D} onto the PCA components with largest explained variance.
        Components 1-6 are shown on the smoothed mid-surface (left) and in unfolded space (right).
        Gray marks missing data.
    }
    \label{fig:method}
\end{figure*}

\textbf{Learning 3D-PLI texture features.}
To learn texture features from \mpli parameter maps, we build on previous work demonstrated for the occipital pole of a vervet monkey brain \cite{oberstrass2024}.
Their approach builds on the observation that brain architecture organizes around spatially consistent groups, which is incorporated in a sampling of spatially close positive pairs in the SimCLR contrastive learning framework \cite{chen2020a} (\cref{fig:method}A).
Here we summarize the approach proposed in \cite{oberstrass2024}.

To obtain positive pairs, consisting of anchor sample $x_a$ and positive sample $x^+$, locations for $x^+$ are drawn from a circle (CL-2D) or sphere (CL-3D) with a fixed radius of \mbox{$r$ = \mmu{118}} around a random sampling location for $x_a$.
The locations are rounded to the next available section, excluding sampling from the same section in CL-3D.
At each sampling location, square image patches of \mpx{128} (\mmu{166}) size are extracted from transmittance $I_T$, direction $\varphi$ and retardation $r$ maps.
On each patch, \mpli specific data transformations $t\sim T$ are applied to become robust to rotation, blur, and color contrast.
The patches are fed to a feature-reduced ResNet-50~\cite{he2016} encoder $f$ to extract 256-dimensional feature vectors $h_a$, $h^+$, which are then projected to 32-dimensional projections $z_a$, $z^+$ by a 2-layer MLP projection head $g$.

For joint training of $f$ and $g$, we create batches of 512 positive pairs.
Per positive pair, samples in all other pairs are used as negative examples for computation of an InfoNCE loss~\cite{oord2018}.
Model training involves all sections of the hippocampus volume except for the 10 most anterior sections, which are kept for validation.
We use Adam optimizer with a learning rate of 10\textsuperscript{-3}, a weight decay of 10\textsuperscript{-6}, \mbox{$\beta_1$ = 0.9}, \mbox{$\beta_2$ = 0}.999 and \mbox{$\epsilon$ = 10\textsuperscript{-8}}.
We set $\tau$ = 0.5 in the InfoNCE loss and train the model to convergence of the validation loss.

After training, projection head $g$ is discarded and deep texture features $h=f(x)$ are extracted for each section in the volume using a sliding window approach with an overlap of 50\% between patches to generate feature maps (\cref{fig:method}B).
By stacking the feature maps, we obtain a new feature activation volume of shape 419 x 545 x 356 with 256 channels.

\textbf{\mpli baseline features.}
To show the effectiveness of deep texture features for representing regional differences in hippocampal architecture, we compare them with baseline features.
First, we adapt fractional anisotropy (FA)~\cite{basser2011} to represent fiber orientations.
To represent the distribution of orientations for patches, we project joint values of $\varphi$ and $\alpha$ onto points on the unit sphere.
The FA values are then calculated from the eigenvalues of their scatter matrix.
As a second feature, we compute mean transmittance values $\bar{I}_T$ for patches of $I_T$.
As transmittance maps $I_T$ show a global similarity with myelin stains~\cite{axer2011} and are affected by the low light transmittance of cell bodies~\cite{zeineh2017} they include information about myelin- and cell body densities.
We compute these baseline features for patches of \mpx{64} (\mmu{83}) to achieve a comparable resolution to the deep texture feature maps.

\textbf{Surface projection and unfolding.}
To analyze the folded architecture of the hippocampus, we apply \textit{HippUnfold}~\cite{dekraker2018}. 
\textit{HippUnfold} provides an unfolded visualization of variations along both the long-axis and proximal-distal axis of the hippocampus~\cite{dekraker2020} as well as a common coordinate system for cross-subject comparison~\cite{dekraker2023}.
Specifically, it generates geometrical inner and outer surfaces of the pyramidal layer of the hippocampal cornu Ammonis (CA) region and the subicular complex~(\cref{fig:method}C).
The surfaces correspond to the geometric inner and outer curvature of the hippocampus and represent the interface between the pyramidal and radiatum layers and between the pyramidal and oriens layers, respectively.

Using surfaces by \textit{HippUnfold}, we sample features from the feature activation volume at several depths, which are generated by interpolating between corresponding vertices of geometrical inner and outer surfaces, including the surfaces themselves.
This enables aggregation of \mpli texture features for the full depth of the pyramidal layer, as demonstrated for cytoarchitectonic features \cite{dekraker2020}.
For each vertex of a mid-surface, we concatenate features at 17 depths into single vectors of size 4\,352.
To reduce dimensionality for visualization and improve computational efficiency in further analysis, we resolve covariance in the features using PCA.
In addition to PCA, we reduce the effect of the cutting angle of the sectioning plane using confound regression \cite{snoek2019}.
The concatenated features are projected onto 52 principal components with largest explained variance (80.1\% total variance explained).
Selected projections are shown in \cref{fig:method}E.
Baseline features are processed in the same way.

\textbf{Evaluation using KMeans clustering.}
We evaluate the distinctiveness of different feature encodings for the architecture of the CA regions by comparing them with subfield labels \cite{dekraker2023} (\cref{fig:clusters}A).
The labels differentiate between \textit{CA1}, \textit{CA2}, \textit{CA3} and \textit{CA4} regions and group areas of the \textit{subicular complex} in a single label.
We introduce an additional label as the \textit{vertical component of the uncus}.
All labels are extrapolated to missing data resulting from above-mentioned truncation.

To determine how well features reflect the regional organization of the hippocampus, we perform k-means clustering for 6 clusters.
Three iterations of graph smoothing by computing mean values over features of neighboring vertices are performed before clustering to reduce noise.
We subsequently compare all clustering results with subfield labels by computing purity, mutual information, and adjusted Rand Index (ARI)~\cite{rand1971}, which measure the homogeneity of labels within clusters, the shared information between clusters and labels, and the similarity between clusters and labels, respectively.

\section{Experiments and Results}
\label{sec:results}

\begin{table}[t!]
  \centering
  \caption{
    Evaluation metrics for k-means clusterings for 6 clusters of features by different feature extraction methods.
    Metrics are averaged over 100 independent runs on 50\% random subsets of vertices and evaluated on the full set.
  }
  \label{tab:cluster_metrics}
  \begin{tabular}{cc|cccc}
    \toprule
      Method & Input & Purity $\uparrow$ & ARI $\uparrow$ & MI $\uparrow$ \\
    \midrule
      \multirow{3}{*}{CL-3D}  & $I_T$, $\varphi$, $r$ & \textbf{0.67} & \textbf{0.33} & \textbf{0.72} \\
      & $I_T$                 & 0.65 & 0.30 & 0.66 \\
      & $\varphi$, $r$        & 0.52 & 0.16 & 0.49 \\
   \midrule
      \multirow{3}{*}{CL-2D}  & $I_T$, $\varphi$, $r$ & 0.63 & 0.26 & 0.61 \\
                              & $I_T$                 & 0.58 & 0.21 & 0.50 \\
                              & $\varphi$, $r$        & 0.51 & 0.13 & 0.40 \\
    \midrule
      $\bar{I}_T$ + FA            & $I_T$, $\varphi$, $r$ & 0.53 & 0.15 & 0.40 \\
      $\bar{I}_T$                   & $I_T$                 & 0.51 & 0.13 & 0.36 \\
      FA                          & $I_T$, $\varphi$, $r$ & 0.50 & 0.11 & 0.27 \\
    \bottomrule
  \end{tabular}
\end{table}

\begin{figure}[t!]
  \centering
  \includegraphics[width=.99\linewidth]{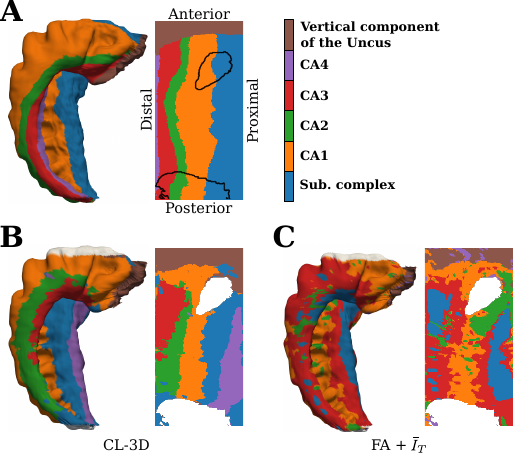}
  \caption{
    Comparison of k-means clustering for 6 clusters with anatomically identified subfield labels (A).
    Clusters of \mbox{CL-3D} texture features (B) more closely resemble the subfield labels than clusters of mean transmittance $\bar{I}_T$ and fractional anisotropy FA (C).
    Images show the midsurface in folded (left) and unfolded space (right).
    Black contours and white spots show positions of missing data.
  }
  \label{fig:clusters}
\end{figure}

We show 6 k-means clusters for \mbox{CL-3D} in \cref{fig:clusters}B and for FA + $\bar{I}_T$ in \cref{fig:clusters}C.
The topographical location and extent of clusters in \mbox{CL-3D} and FA + $\bar{I}_T$ features show a good agreement with that of hippocampal CA1 - CA4 regions and the subicular complex.
Compared to FA + $\bar{I}_T$, \mbox{CL-3D} features show a clearer alignment and reduced noise in the clusters.

For a quantitative comparison of k-means clusters with subfield labels, we average metrics across 100 independent initializations of k-means, each fitted to random subsets of 50\% of vertices and evaluated on the entire dataset (\cref{tab:cluster_metrics}).
Given the same input modalities, deep texture features consistently achieve higher scores than the baseline features, with \mbox{CL-3D} outperforming \mbox{CL-2D}.
While the highest scores for each method are achieved when using all input modalities \mbox{($I_T$, $\varphi$, $r$)}, features derived from transmittance maps alone perform almost as well.
Features based only on fiber orientation information encoded in $\varphi$ and $r$ achieve lower scores compared to features based on transmittance values $I_T$.

\section{Discussion \& Conclusion}
\label{sec:discussion}
In this study, we combined deep \mpli texture features with geometric unfolding to derive a novel approach for characterizing the regional organization of human hippocampal CA regions.
The texture features were extracted using a recently proposed, fully data-driven representation learning approach for \mpli \cite{oberstrass2024}, and evaluated in a canonical surface space obtained via topological unfolding \cite{dekraker2018,dekraker2020}.
Although the features do not cluster into a precise delineation of hippocampal subfields as identified by classical cytoarchitectonic criteria, they follow the general regional organization pattern and additionally highlight an expected functional rostro-caudal heterogeneity without explicit prior information on anatomy.

\vspace*{\fill}

Our experiments have shown that features extracted from transmittance maps alone already form clusters that correspond well to hippocampal subfields.
This is expected, as transmittance maps contain information about cyto- and fiber architecture \cite{zeineh2017}.
Including fiber orientation information as input for feature extraction, however, further improves correspondence with subfields.
This is in line with neuroanatomical studies that consider myeloarchitecture for subfield delineation \cite{ding2015}.
We have further shown that cross-section sampling of positive pairs in contrastive learning leads to better clustering of \mpli texture by subfields compared to in-plane sampling, which are both considered in \cite{oberstrass2024}.

\vspace*{\fill}

While this study focused on a single specimen, the approach is fully unsupervised and can thus be applied to other brains, specimens, and microscopic modalities with available 3D reconstructions.
This will be part of future investigations.
Projecting deep texture features to unfolded space using \textit{HippUnfold} \cite{dekraker2018} enables subsequent correlation with diverse modalities \cite{dekraker2023}.
The presented work thus lays the foundation for incorporating \mpli texture information into a comprehensive multimodal mapping of the human hippocampus.

\clearpage

\section{Compliance with ethical standards}
\label{sec:ethics}
The postmortem brain for this study was obtained through the body donor program of the University of Rostock, Germany, and in accordance with the local ethics committee.
The present study does not require additional ethical approval.
All body donors have signed a declaration of agreement.

\section{Acknowledgments}
\label{sec:acknowledgments}
This project received funding from the Helmholtz Association’s Initiative and Networking Fund through the Helmholtz International BigBrain Analytics and Learning Laboratory (HIBALL), Helmholtz International Lab grant agreement InterLabs-0015, and the European Union’s Horizon 2020 Research and Innovation Programme, grant agreement 945539 (HBP SGA3), which is now continued in the European Union’s Horizon Europe Programme, grant agreement 101147319 (EBRAINS 2.0 Project).
Computing time was granted through JARA on the supercomputer JURECA at Jülich Supercomputing Centre (JSC).

\bibliographystyle{bib/IEEEbib}

\bibliography{bib/refs_manual.bib}

\clearpage

\end{document}